\documentclass[a4paper, 10pt, conference]{ieeeconf}      
\usepackage{FG2021}

\FGfinalcopy 

\usepackage{times}
\usepackage{epsfig}
\usepackage{graphicx}
\usepackage{amsmath}
\usepackage{amssymb}
\usepackage{comment}
\usepackage{subfig}
\usepackage{booktabs}
\usepackage{adjustbox}
\usepackage{hyperref}

\newcommand{\etal}{\textit{et al.~}}

\usepackage{array}
\usepackage{booktabs}
\usepackage{multirow}

\newcolumntype{R}[2]{%
    >{\adjustbox{angle=#1,lap=\width-(#2)}\bgroup}%
    l%
    <{\egroup}%
}

\IEEEoverridecommandlockouts                              

\def\FGPaperID{57} 

\title{\LARGE \bf
Self-Supervised 3D Human Pose Estimation\\ with Multiple-View Geometry
}

\author{\parbox{16cm}{\centering
    {\large Arij Bouazizi$^{1,2}$, Julian Wiederer$^{1,2}$, Ulrich Kressel$^1$ and Vasileios Belagiannis$^2$}\\
    {\normalsize
        $^1$ Mercedes-Benz AG, Stuttgart, Germany \\
    $^2$ Universit\"at Ulm, Ulm, Germany.
}}
    \thanks{E-mail: \textit{firstname.lastname@\{daimler.com, uni-ulm.de\}}.}
}

\begin{document}

\ifFGfinal
\thispagestyle{empty}
\pagestyle{empty}
\else
\author{Anonymous FG2021 submission\\ Paper ID \FGPaperID \\}
\pagestyle{plain}
\fi
\maketitle

\begin{abstract}

We present a self-supervised learning algorithm for 3D human pose estimation of a single person based on a multiple-view camera system and 2D body pose estimates for each view. To train our model, represented by a deep neural network, we propose a four-loss function learning algorithm, which does not require any 2D or 3D body pose ground-truth. The proposed loss functions make use of the multiple-view geometry to reconstruct 3D body pose estimates and impose body pose constraints across the camera views. Our approach utilizes all available camera views during training, while the inference is single-view. In our evaluations, we show promising performance on Human3.6M and HumanEva benchmarks, while we also present a generalization study on MPI-INF-3DHP dataset, as well as several ablation results. Overall, we outperform all self-supervised learning methods and reach comparable results to supervised and weakly-supervised learning approaches. Our code and models are publicly available\footnote{Source Code: \url{https://github.com/vru2020/Pose_3D/}}.
\end{abstract}

\section{Introduction}

Image-based human pose estimation experienced an enormous improvement the past few years thanks to deep neural networks and large-scale annotated databases. In particular, 2D human pose estimation approaches showed excellent results in almost any kind of image context~\cite{newell2016stacked, papandreou2018personlab}. Unlike, 3D human pose estimation is not at the same level due to the difficulty of obtaining ground-truth information and the higher complexity of the mapping to the three-dimensional space. Multi-camera systems and motion capture sensors are not massively available while setting up such a system is much more complicated than annotating 2D body postures in images.

\begin{figure}[ht]
    \centering

    \includegraphics[width=0.45\textwidth, height=0.45\textwidth]{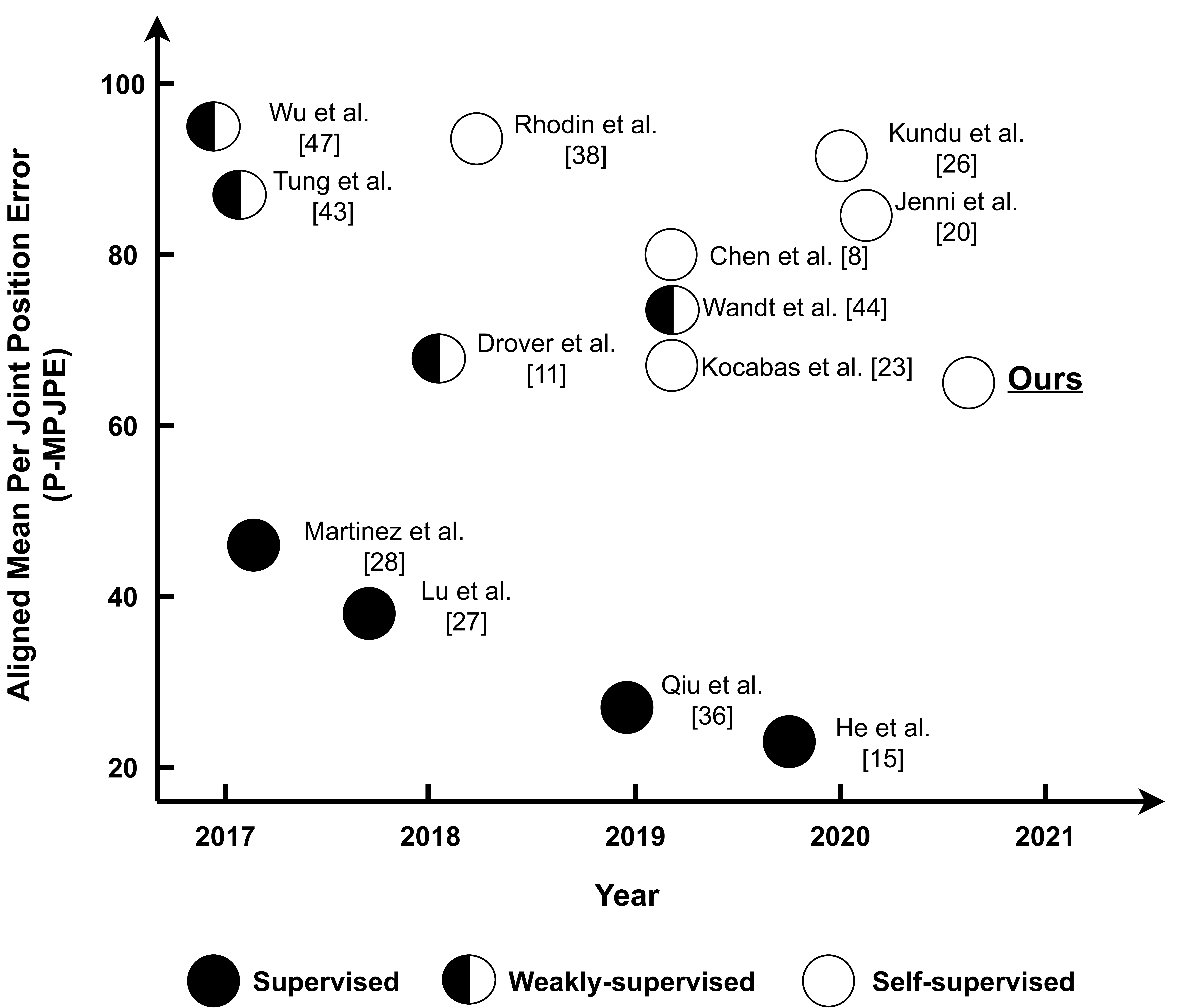}

    \caption{\textbf{Human3.6M~\cite{ionescu2013human3} Results.} We present a  self-supervised learning approach for single human 3D body pose estimation. Compared to supervised and weakly-supervised approaches, we deliver promising results by relying on pseudo-labels and four-loss functions for self-supervision.} 
    \label{fig:teaser}
\end{figure}

In this work, we address the problem of 3D human pose estimation for a single person without demanding 2D or 3D body pose ground-truth information. Our supervision comes from 2D body pose estimates and the geometry properties of a multiple-view calibrated camera system. We propose a self-supervised learning-based algorithm to map a 2D body pose estimate, coming from an image, to the 3D body pose in the camera coordinate system. Our learning algorithm makes use of all available camera views in the training, while the inference is single-view.

Learning-based approaches for 3D human pose estimation still reach the best results with supervised learning~\cite{he2020epipolar, Martinez17, pavlakos2017coarse, tu2020voxel}. However, other types of supervision have been recently explored to minimize the process of 3D body pose annotation. Weakly- and semi-supervised learning methods make use of unpaired 2D -- 3D body pose data~\cite{tung2017adversarial, wandt2019repnet} or rely on a small set of annotated 3D body pose data to fine-tune a generic model~\cite{Rohdin2018}. Lately, self-supervised learning has attracted much attention because of the independence from ground-truth information~\cite{jakab2020self, novotny2019c3dpo, tripathi2020posenet3d}. In these approaches, adversarial training often takes care of the body pose geometry learning task. In our approach, we follow the idea of self-supervision, but instead, rely on a multiple-view geometry to learn the body pose constraints.

We propose an algorithm with four-loss functions for learning to predict the 3D body pose from 2D body pose estimate input. We first generate 3D body pose estimates by triangulating the 2D pose estimates. The 3D estimates (pseudo-labels) serve as supervision for training our deep neural network. Second, we introduce the re-projection loss for self-supervision that minimizes the difference between the 2D body pose projections of the 3D predictions and 2D body pose input estimates. To obtain extra supervision and impose geometric constraints, we project the 3D body pose prediction to all camera views and minimize the difference to the 2D body pose estimates. Third, we impose body pose geometric constraints across all camera views by forcing the 3D body pose predictions from all views to be the same. Finally, we propose to project the 3D body predictions and triangulate again to obtain additional 3D body pose estimates (considered as ground-truth too). We show in practice that triangulating the projected 3D body pose predictions results in improving the final performance. The motivation for this loss function is to be independent of the input 2D estimates, which contribute to the rest of the loss functions. In our experiments, we rely on two standard benchmarks to evaluate the four proposed loss functions for self-supervision. First, we show state-of-the-art performance on Human3.6M~\cite{ionescu2013human3} dataset and then on HumanEva~\cite{sigal2010humaneva} dataset. In addition, we present an ablation study and a transfer learning evaluation on MPI-INF-3DHP~\cite {mpiiinfdataset} to further examine the generalization and analyze our approach.

To sum up, our work makes the following contributions: 1) we rely only on 2D body pose estimates and multiple-view calibrated system to obtain supervision for monocular 3D human pose estimation -- we make thus use of pseudo-labels instead of 3D ground-truth information, 2) we propose a model agnostic self-supervised learning algorithm with four-loss functions, 3) we present an extensive evaluation using standard benchmarks where we reach the best results in the field for learning without 3D ground-truth approaches and comparable results to supervised and weakly-supervised methods.


\section{Related Work}

Human body pose estimation is a long-standing problem in computer vision. Below, we present the relevant prior work on learning-based 3D human pose estimation.

\paragraph{Supervised Learning}

In the past, learning-based approaches normally relied on body part detectors and multiple-view geometry~\cite{Belagiannis2014, belagiannis20153d, joo2015panoptic}. With the deep learning revolution, ConvNet-based detectors have significantly improved 3D human pose estimation~\cite{huang2017towards}, while at the same time the multi-view information utilization became less necessary. Single-view 3D human pose estimation showed comparable and often better results than multi-view approaches, using deep neural networks~\cite{pavlakos2017coarse, Martinez17, moreno20173d}. Nevertheless, multiple-view geometric priors still lead to state-of-the-art performance~\cite{qiu2019cross, he2020epipolar, tu2020voxel}. Despite the remarkable results, these approaches require ground-truth data to be trained. In this work, we show how to reach similar results without relying on 3D ground-truth information.

\paragraph{Weakly Supervised Learning}
Weakly supervised learning relaxed the assumption of annotating all data. The annotation stems from fitting a given 3D human model to the image data~\cite{zanfir2020weakly} or unpaired 2D and 3D body pose annotations~\cite{tung2017adversarial, drover20183d, kanazawa2018end, wandt2019repnet}. Using unpaired 2D and 3D pose data is usually combined with adversarial learning. For instance, Wandt \textit{et.~al.}~\cite{wandt2019repnet} propose the re-projection network to learn the mapping from 2D to the 3D body pose distribution using adversarial learning. In particular, the critic network improves the generated 3D body pose estimate based on the Wasserstein loss~\cite{arjovsky2017wasserstein} and unpaired 2D and 3D body poses. Similarly, Drover~\textit{et.~al.}~\cite{drover20183d} rely on a discriminator network for supervision of 2D body pose projections. However, the method additionally utilizes 3D ground-truth data to synthetic 2D body joints for training. Instead of adversarial learning, we rely on a multiple-view calibrated system to reach the same goal. In addition, our approach does not demand any kind of 3D ground-truth body poses.

\paragraph{Self-Supervised Learning without 3D ground-truth}
 Self-supervised Learning has recently shown promising 3D body pose results thanks to the robust 2D pose estimation algorithms~\cite{newell2016stacked, papandreou2018personlab} and the self-supervision loss functions~\cite{novotny2019c3dpo, jakab2020self}. The self-supervision usually comes from the multiple-view geometry~\cite{Kocabas19} or the video constraints~\cite{kundu2020selfsupervised}. Adversarial learning offers self-supervision too~\cite{drover20183d, kudo2018unsupervised}. Kocabas \textit{et.~al.}~\cite {Kocabas19}, triangulate 2D pose estimate in multi-view environment to generate pseudo-labels for 3D body pose training. Chen \textit{et.~al.}~\cite{chenunsupervised19} propose the self-consistency loss and adversarial training for
 self-supervised training. The self-consistency loss is motivated by the fact that predicted 3D skeletons can be randomly rotated and projected without any change in the distribution of the resulting 2D skeletons. In our work, we explore both ideas of pose estimate triangulation and 3D body pose consistency for a multiple-view system. Since the pose estimates can be noisy, we additionally propose another self-supervised loss that does not depend on the pose estimate. Tripathi \textit{et.~al.}~\cite{tripathi2020posenet3d} propose a knowledge distillation approach to compensate for the lack of ground-truth. The approach also relies on dilated convolutions to model the temporal dynamics which act as extra self-supervision. Lately, Kundu \textit{et.~al.}~\cite {kundu2020selfsupervised} present a self-supervised 3D pose estimation method with an interpretable latent space that allow view synthesis in addition to 3D human pose estimation and body part segmentation. The approach relies though on unpaired 3D annotation to improve the final result. Rhodin \textit{et.~al.}~\cite{Rohdin2018} rely on a multi-view configuration to learn a latent representation without supervision. However, some annotated samples are required to fine-tune the model on a semi-supervised manner. Unlike, we require neither unpaired 3D poses nor 3D ground-truth information to train our approach. Instead, we generate 3D pseudo-labels.

\section{Method}

In this section, we present our learning algorithm for 3D human body pose estimation from 2D body pose estimates. The supervision stems from the multi-view geometry properties of a calibrated system and the 2D pose detection from a body pose estimation algorithm. We present below how we train a deep neural network based on our proposed loss functions.

\subsection{Problem definition}

Consider a multi-view system with $C$ cameras, which is synchronized in time and calibrated. For each camera view $c$, a set of 2D body pose estimates $\mathbf{S}_{c} = \{\mathbf{\hat{y}}^{s}_{c}\}^{|\mathbf{S}_{c}|}_{s=1}$, is available using a pose estimation algorithm. A 2D body pose estimate $\mathbf{\hat{y}}^{s}_{c}$ corresponds to a tuple of $N$ landmarks in the image plane, such that $\mathbf{\hat{y}}^{s}_{c}= (\mathbf{\hat{y}}^{s}_{c,1}, \dots, \mathbf{\hat{y}}^{s}_{c,N})$. For the human pose estimation problem, an estimated landmark $\mathbf{\hat{y}}^{s}_{c,n} = (x^{s}_{c,n}, y^{s}_{c,n})$ usually corresponds to a body joint. To obtain the estimates, we can rely on some 2D human pose estimation algorithm~\cite{belagiannis2014holistic, newell2016stacked, papandreou2018personlab}.

Given the calibrated system and the 2D body pose estimates only, our goal is to learn the mapping from the 2D body pose $\mathbf{y}$ to the 3D body pose $\mathbf{Y}$ with a deep neural network $f_{\mathbf{\theta}} (\cdot)$, parametrized by $\mathbf{\theta}$. The network parameters $\mathbf{\theta}$ are learned in a self-supervised way with four-loss functions that rely on the 2D pose estimates across all views and 3D pose estimates, coming from network input triangulation and the triangulation of the projected network output. Note that our approach targets single person 3D body pose estimation.
\subsection{Learning Algorithm}

\begin{figure*}[t]
    \centering
    \includegraphics[width=\textwidth]{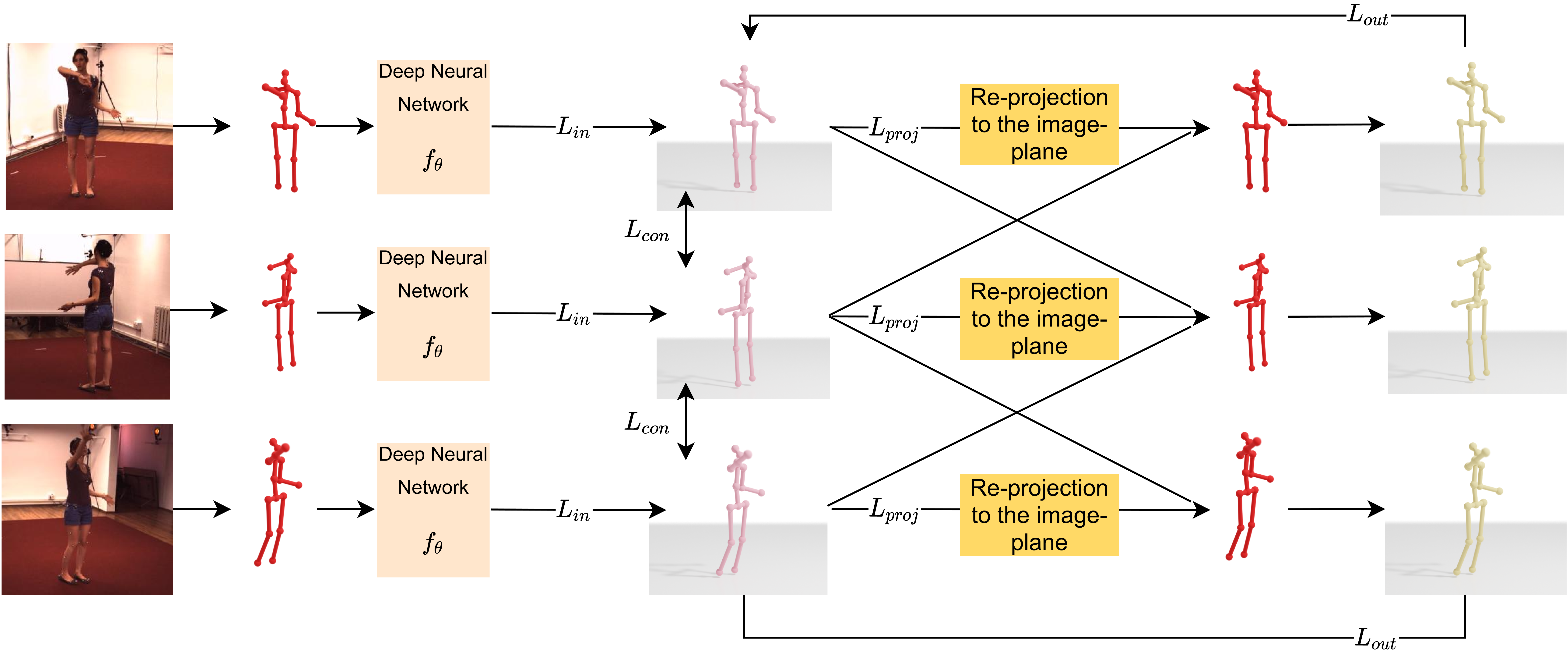} 
    \caption{Illustration of our learning algorithm.  During training we optimize the deep neural network $f_{\mathbf{\theta}} (\cdot)$ to map a 2D pose estimate to 3D body pose with four loss functions. As input to our model we assume 2D pose detections from multiple views shown on the left. As first geometric self-supervision, we compute the triangulated 3D pose from the 2D detections as estimated ground-truth for the input triangulation loss $\mathcal{L}_{in}$. For 3D consistency between the 3D body pose predictions from multiple views, we define the consistency loss $\mathcal{L}_{con}$. We re-project the 3D poses estimated by the network to the image plane to define two extra loss functions. The output triangulation loss $\mathcal{L}_{out}$ introduces an additional self-supervision path by triangulating the re-projected 3D predictions. For consistency in 2D the re-projected predictions are evaluated in a re-projection loss $\mathcal{L}_{proj}$.}
    \label{fig:method}
\end{figure*}

Our learning algorithm is illustrated in Fig~\ref{fig:method}. Given the training sample $s$, i.e.~a set of $C$ images, we first apply the human pose detector on each view to extract 2D body pose estimates, which serve as input to $f_{\mathbf{\theta}}(\cdot)$, and for obtaining the 3D human pose estimate $\mathbf{\hat{Y}}^{s}_{in}$ by triangulating based on all camera views, where $\mathbf{\hat{Y}}^{s}_{in} = (\mathbf{\hat{Y}}^{s}_{in,1}, \dots, \mathbf{\hat{Y}}^{s}_{in, N})$ and $\mathbf{\hat{Y}}^{s}_{in, n} = (X^{s}_{in,n}, Y^{s}_{in,n}, Z^{s}_{in,n})$ are 3D coordinates in the camera coordinate system. The input triangulation serves as supervision for the \textit{input triangulation loss} $\mathcal{L}_{in}$, which minimizes the difference of the 3D body pose prediction $\mathbf{\widetilde{Y}}^{s}$ and triangulated 3D body pose $\mathbf{\hat{Y}}^{s}_{in}$ (set as ground-truth). In addition, we introduce our second loss function which is applied on each camera view. In detail, the \textit{re-projection loss} $\mathcal{L}_{proj}$ adds self-supervision by minimizing the difference between the input of each camera view $\mathbf{\hat{y}}^{s}_{c}$ (taken as ground-truth) and the projection of the 3D body pose prediction $\mathbf{\widetilde{y}}^{s}_{c}$ to the camera view $c$. Third, we force consistency between the 3D body pose predictions across all views. For this reason, we present the \textit{consistency loss} $\mathcal{L}_{con}$ that considers all pairs of views and minimizes the difference between the predicted 3D body poses. Fourth, we propose to perform triangulation of the projected 3D body pose prediction across all views. We refer to this triangulation as output triangulation and consider it as self-supervision to train the model. For this reason, we design the \textit{output triangulation loss} $\mathcal{L}_{out}$ that minimizes the difference between the output triangulation $\mathbf{\hat{Y}}^{s}_{out}$ (taken as ground-truth) and the predicted 3D body pose $\mathbf{\widetilde{Y}^{s}}$ given the sample $s$. This loss function applies to all camera views as well.

To learn the parameters $\mathbf{\theta}$, we train our model based on the proposed loss functions and using the training samples from all camera views $\{\mathbf{S}_{c}\}^C_{c=1}$. We obtain the model parameters by minimizing the following objective: 
\begin{equation}
     \theta^{\prime} = \arg \min_{\theta} \omega_{1}\mathcal{L}_{in} + \omega_{2} \mathcal{L}_{proj} + \omega_{3}\mathcal{L}_{con}  + \omega_{4}\mathcal{L}_{out},
\end{equation} 
where $\omega_{1}$, $\omega_{2}$,  $\omega_{3}$, and $\omega_{4}$ are weighing factors for each loss. Next, we describe in detail each loss function.

\subsection{Loss Functions}
\label{loss_func_sec}

We present the motivation and elements of each loss function below.

\paragraph{Input Triangulation Loss}
To compute the 3D position for each landmark based on the camera view detections, we make use of the Direct Linear Triangulation (DLT) method~\cite {Cambridge2004}, which is differentiable and thus suits-well to our optimization. The same transformation applies to all body pose landmarks. Similar to \cite{Kocabas19}, we consider the obtained 3D human pose estimate $\mathbf{\hat{Y}}^{s}_{in}$ as ground-truth. We define the input triangulation loss as:
\begin{equation}\label{input_loss}
\mathcal{L}_{in} = \sum_{s=1}^{S} \sum_{c=1}^{C} \parallel \rho_{w \rightarrow c}(\mathbf{\hat{Y}}^{s}_{in})  - f_{\mathbf{\theta}}(\mathbf{\hat{y}}^{s}_{c}) \parallel^2,
\end{equation}
where $\rho_{w \rightarrow c} (\cdot)$ corresponds to the transformation from the world coordinate system $w$ to the camera $c$ coordinate system. The loss depends on the quality of the detected landmarks for each camera view. We further discuss this point in the experiments.

\paragraph{Re-Projection Loss}
The re-projection loss~\cite{tung2017adversarial} provides multiple-view supervision, which mitigates the ambiguities and occlusions that arise from the single view 2D detections. For each camera view $c$, the predicted 3D body pose is projected to all other camera views and to the current one. Then, the difference between the input 2D pose (considered as ground-truth) and all projections is minimized. We define the re-projection loss as:
\begin{equation}\label{proj_loss}
\mathcal{L}_{proj} = \sum_{s=1}^{S} \sum_{c=1}^{C} \sum_{c\prime=1}^{C} {\parallel \mathbf{\hat{y}}^{s}_{c} - \tau_{c}(f_{\mathbf{\theta}}(\mathbf{\hat{y}}^{s}_{c\prime})) \parallel },
\end{equation}
where $\tau_{c}$ corresponds to the projection of the predicted 3D pose to the camera view $c$. We choose the L1 loss for the minimization since we observed faster convergence in our experiments. The double summation in Eq.~\ref{proj_loss} shows that we project the predicted 3D body pose across all camera views. We empirically found that the additional operations, compared to single camera projection, have a positive impact on the model performance. Similar to the input triangulation loss (Eq.~\ref{input_loss}), the re-projection loss is depended on the input 2D pose detection quality. However, we make the loss more robust to false detections by considering all camera views.

\paragraph{Consistency Loss}
We impose geometric body pose constraints in the 3D predictions with the consistency loss. In principle, the 3D body pose prediction from each camera view should be the same given the training sample $s$, regardless the 2D body pose input. The only difference is the coordinate system in which the 3D body pose is expressed. Based on this observation, we propose the consistency loss that is given by: 
\begin{equation}\label{cons_loss}
\mathcal{L}_{con} = \sum_{s=1}^{S} \sum_{c=1}^{C} \sum_{\substack{c\prime=1 \\ c \neq c\prime}}^{C} {\parallel f_{\mathbf{\theta}}(\mathbf{\hat{y}}^{s}_{c}) - \rho_{c\prime \rightarrow c}(f_{\mathbf{\theta}}(\mathbf{\hat{y}}^{s}_{c\prime})) \parallel }
\end{equation}
where $\rho_{c\prime \rightarrow c} (\cdot)$ corresponds to the transformation from the camera $c\prime$ to the camera $c$ coordinate system. The idea is to transform the 3D prediction from the camera view $c\prime$ to all other cameras and assure that the predictions looks the same. Note that we consider the camera view $c$ as the ground-truth to minimize the difference from the transformed 3D prediction. Similar to the re-projection loss, we empirically select the L1 loss for the minimization due to the fast convergence.

\paragraph{Output Triangulation Loss}

The three presented loss functions are directly influenced by the quality of the 2D input estimates. To address this point, we propose the output triangulation loss that is independent of the 2D pose detection input. For the output triangulation loss, we first create a new 3D body pose estimate (as ground-truth) by projecting the predicted 3D body pose from each camera view to the image plane and use the $C$ body pose projections to triangulate again. This triangulated 3D body pose $\mathbf{\widetilde{Y}}_{out}$, which is based on the projected predictions, serves as ground-truth for the output triangulation loss defined as:
\begin{equation}\label{output_loss}
\mathcal{L}_{out} = \sum_{s=1}^{S} \sum_{c=1}^{C} \parallel \rho_{w \rightarrow c}(\mathbf{\widetilde{Y}}^{s}_{out})  - f_{\mathbf{\theta}}(\mathbf{\hat{y}}^{s}_{c}) \parallel^2.
\end{equation}
where $\rho_{w \rightarrow c} (\cdot)$ corresponds to the transformation from the world coordinate system $w$ to the camera $c$ coordinate system. In practice, we observed that this loss helps to the model convergence at the last stage of training where the predictions are meaningful. For this reason, we combine it with the rest loss functions once the model makes reasonable 3D body pose predictions. Since we rely on a single network architecture for all experiments, this rule works well for all of our evaluations.

\subsection{Network Architecture} 

In our problem, it is common to design fully connected network architectures to estimate the 3D keypoints. For that reason, we implement the model of Martinez \textit{et al.}~\cite {Martinez17}. The encoder takes $N$ 2D landmarks as input, e.g.~16, represented by a vector of $2N$ elements, to a fully connected layer with 1024 output channels with four subsequent residual blocks. A residual block is composed of a pair of fully connected layers, each with 1024 neurons followed by batch normalization and ReLu activation function. The decoder has the same architecture as the encoder computing an output vector with $3N$ dimensions representing $N$ 3D joint locations.

\subsection{Single-View Inference} 
In training, our learning algorithm is based on a calibrated camera-system. Nevertheless, we could skip the calibration and extract the camera parameters as in \cite{Kocabas19}.

During inference, our approach works with the 2D pose estimate from a single view to make 3D body pose predictions given by:
\begin{equation}\label{single_view_inference}
    \mathbf{\widetilde{Y}} = f_{\mathbf{\theta}} (\mathbf{\hat{y}}).
\end{equation}
Next, we present the evaluation of our learning algorithm.

\section{Experiments}

We evaluate our approach on two standard benchmarks for 3D human pose estimation, namely Human3.6M~\cite{ionescu2013human3} and HumanEva~\cite{sigal2010humaneva} datasets. Furthermore, we examine the contribution of each loss function in the ablation study. At last, we determine the generalization of our approach by applying a pre-trained model from Human3.6M~\cite{ionescu2013human3} to the MPII-INF-3DHP~\cite{mpiiinfdataset} dataset. 

\subsection{Implementation details}

\paragraph{Body Pose Representation} Similar to the prior work~\cite{Martinez17,drover20183d,chen20173d,Kocabas19}, we make 3D predictions on the camera coordinate system where we set the origin of the coordinate system at the \textit{pelvis} joint. We set the same origin for the image plane too.

\paragraph{2D Body Pose Detection}
The 2D body pose estimates come from pose detectors, which we select according to the dataset. In the Human3.6M~\cite{ionescu2013human3} evaluation, we use the same 2D detections as~\cite{pavllo20193d}, which are obtained after running Mask R-CNN~\cite{he2017mask} to all camera views and obtaining the 2D body pose with the fine-tuned Cascade Pyramid Network~\cite{chen2018cascaded}. In HumanEva and MPII-INF-3DHP, we first rely on Mask R-CNN to detect the subject in the image plane. Then, we apply the Stacked Hourglass model~\cite{newell2016stacked} pre-trained on the MPII Human Pose dataset~\cite{andriluka20142d} to extract the 2D body pose. Finally, we follow the prior work to down-sample all the videos from 50 to 10 frames per second~\cite{Martinez17, pavllo20193d, Kocabas19}.

\begin{table}[t]
    \caption{Results on the Human3.6M dataset. We show the P-MPJPE on the test set of Human3.6M for models with different levels of supervision. The best performing method is marked in bold. In comparison with all current state-of-the art self-supervised and weakly supervised methods, our approach performs best.}
	\label{tab:h36m_aligned_mpjpe_short}
	\centering
	\setlength{\tabcolsep}{2pt}
	\begin{adjustbox}{max width=0.8\linewidth}
		\begin{tabular}{l l l l | l} 
			\toprule
			Supervision&&& Method &  P-MPJPE \\
			\midrule
			\multirow{5}{*} {\rotatebox{90}{Supervised}}
			&&& Akther and Black ~\cite{akhter2015pose}  &  181.1   \\ 
			&&& Zhou et al. ~\cite {zhou2016sparse}  & 106.7  \\
			&&& Bogo et al. ~\cite {bogo2016smpl}  &  82.3  \\
			&&& Martinez et al. ~\cite {Martinez17}  &  47.7 \\
			&&& Lu et al. ~\cite {Lu2018}  &  46.6 \\
			\midrule   
			\multirow{4}{*}{\rotatebox{90}{Weakly}} \multirow{4}{*}{\rotatebox{90}{supervised}}
		    &&& Wu et al. ~\cite {Wu_2016}  &  98.4 \\
			&&& AIGN. ~\cite{tung2017adversarial} &  97.2 \\
			&&& Wandt et al. ~\cite {wandt2019repnet}  &  65.1 \\
			&&& Drover et al. ~\cite {drover20183d}  &  64.6 \\
			\midrule 
			\multirow{6}{*}{\rotatebox{90}{Self}} \multirow{6}{*}{\rotatebox{90}{supervised}}
			&&& Rhodin et al. \cite {Rohdin2018}  &  98.4 \\
			&&& Kundu et al. \cite{kundu2020selfsupervised} &  85.8 \\
			&&& Jenni et al. \cite {jenni2020selfsupervised} & 78.4 \\
			&&& Chen et al. \cite{chenunsupervised19}  &68.0 \\
			&&& Kocabas et al. \cite {Kocabas19}  & 67.5 \\

			\midrule 
			&&&  Ours & \textbf{62.0} \\
			\bottomrule
		\end{tabular}%
	\end{adjustbox}
\end{table}

\begin{table*}[h!]
  \begin{center}
  \caption{Results on the Human3.6M dataset. Comparison of our self-supervised approach with state-of-the-art supervised and weakly supervised methods following evaluation Protocol-II (with rigid alignment) individually for all 15 actions.}
     \label{tab:h36m_aligned_mpjpe}
    \resizebox{\textwidth}{!}{\begin{tabular}{l l l | c c c c c c c c c c c c c c c | c}
    \toprule
       &&Method & Dir. & Dis. & Eat & Greet & Phone & Photo & Pose
                          & Purch. & Sit & SitD & Smoke & Wait & WalkD & Walk
                          & WalkT & Avg\\
      \hline
      \multirow{5}{*} {\rotatebox{90}{Sueprvised}} && Akther \& Black \cite {akhter2015pose} & 199.2 & 177.6 & 161.8 & 197.8 & 176.2  & 186.5 & 195.4 & 167.3 & 160.7 & 173.7 &  177.8 & 181.9 & 198.6 & 176.2 & 192.7 & 181.1 \\
      && Zhou et al. \cite {zhou2016sparse} & 99.7 & 95.8 & 87.9 & 116.8 & 108.3  & 107.3 & 93.5 & 95.3 & 109.1 & 137.5 &  106.0 & 102.2 & 110.4 & 106.5 & 115.2 & 106.7 \\
      && Bogo et al. \cite {bogo2016smpl} & 62.0 &  60.2 & 67.8 & 76.5 & 92.1  & 77.0 & 73.0 & 75.3 & 100.3 & 137.3 &  83.4 & 77.3 & 79.7 & 48.0 & 87.7 & 82.3 \\
      && Martinez et al. \cite {Martinez17} & 39.5 &  43.2 & 46.4 & 47.0 & 51.0  & 56.0 & 41.4 & 40.6 & 56.5 & 69.4 & 49.2 & 45.0 & 38.0 & 49.0 & 43.1 & 47.7 \\
      && Lu et al. \cite {Lu2018} & 40.8 &  44.6 & 42.1 & 45.1 & 48.3  & 54.6 & 41.2 & 42.9 & 55.5 & 69.9 &  46.7 & 42.5 & 36.0 & 48.0 & 41.4 & 46.6 \\[0.3em]
      \hline
      \multirow{5}{*} {\rotatebox{90}{Weakly}} \multirow{5}{*} {\rotatebox{90}{supervised}} && Wu et al. \cite {Wu_2016} & 78.6 &  90.8 & 92.5 & 89.4 & 108.9 & 112.4 & 77.1 & 106.7 & 127.4 & 139.0 & 103.4 & 91.4 & 79.1 & - & - & 98.4 \\
      && AIGN. \cite{tung2017adversarial} & 77.6 &  91.4 & 89.9 & 88.0 & 107.3 & 110.1 & 75.9 & 107.5 & 124.2 & 137.8 &  102.2 & 90.3 & 78.6  & - & - & 97.2\\  
      && Wandt et al. \cite {wandt2019repnet} & 53.0 &  58.3 & 59.6 & 66.5 & 72.8 & 71.0 & 56.7 & 69.6 & 78.3 & 95.2 & 66.6 & 58.5 & 63.2 & 57.5 & 49.9 & 65.1 \\
      && Drover et al. \cite {drover20183d} & 60.2 &  60.7 & 59.2 & 65.1 & 65.5 & 63.8 & 59.4 & 59.4
                      & 69.1 & 88.0 &  64.8 & 60.8 & 64.9  & 63.9 & 65.2 & 64.6\\[0.3em]
      \hline 
      && Ours (self-supervised)  & 49.4 &  51.7 & 61.7 & 56.5 & 64.9 & 67.1 & 51.6
                      & 52.1 & 83.9 & 111.3 & 60.5 & 54.7 & 56.9 & 45.9 & 53.6 & \textbf{62.0} \\
      \bottomrule
    \end{tabular}}
  \end{center}
\end{table*}

\paragraph{Training}
For the Human3.6M training, we have a batch size of $8192$, learning rate $1e-3$, $500$ training epochs and rely on the Adam optimizer. For the HumanEva training, we use the same optimizer for 5000 epochs with a learning rate of $1e-3$ and a batch size of $1600$. Since some frames are corrupted by sensor dropout, we skip them and only use the valid frames for training. Finally, we empirically found that the output triangulation loss works at best when used at the last $10\%$ of training epochs. The weighting factors for each loss are $\omega_{1} = 1$, $\omega_{2} = 1$,  $\omega_{3} = 0.001$, and $\omega_{4} = 0.01$, found empirically. We set the weighting factors fixed for all experiments.

\subsection{Human3.6M Evaluation}
Human3.6M~\cite{ionescu2013human3} is the de facto benchmark for 3D human pose estimation. In the dataset 7 professional actors perform 15 different daily actions, including discussion, eating, sitting or walking motions. The dataset contains 3.6 million 3D human body poses, generated based on a motion capture system and four cameras. Similar to the related work~\cite{chenunsupervised19,kundu2020selfsupervised,Martinez17,tung2017adversarial}, our metric is the mean per joint position error after Procrustes alignment (P-MPJPE). We follow the standard protocol using the five actors 1, 5, 6, 7 and 8 for training and the remaining two actors 9 and 11 for testing. We  summarize  our results in Table~\ref{tab:h36m_aligned_mpjpe_short}. In Table~\ref{tab:h36m_aligned_mpjpe}, we additionally provide the results per action and compare with approaches that include the same evaluation. We outperform all related work for self-supervised learning without 3D ground-truth information. Besides, we outperform all weakly supervised learning approaches, as well as some of the supervised learning too. Similar to us, the approaches of~\cite{Rohdin2018,Kocabas19} leverage multiple-view information for 3D human pose estimation. In particular, we have a common loss function with~\cite{Kocabas19}, namely the input triangulation loss. Nevertheless, we clearly show better performance than both approaches thanks to our four-loss function formulation. We demonstrate some visual results in Fig.~\ref{subfig:h36m_qualitative_results}.

\subsection{HumanEva Evaluation}
\begin{figure*}[t]
\centering
\begingroup
\captionsetup[subfigure]{width=\textwidth}

\subfloat[Qualitative results on the Human3.6M dataset. The model is able to reconstruct the 3D body pose from a single 2D pose detection in various actions like phoning, eating or discussing.]{
	\label{subfig:h36m_qualitative_results}
	\includegraphics[width=\textwidth]{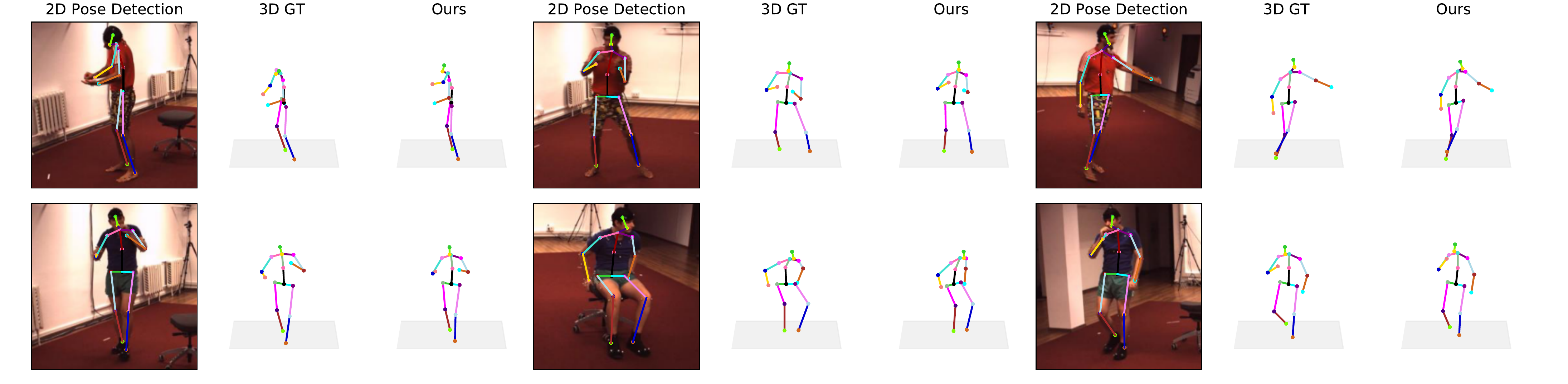}}
 
\subfloat[Qualitative results on the HumanEva dataset. The test set of HumanEva includes three actions, walking, jogging and boxing. In all actions, the model shows results close to the ground-truth.]{
	\label{subfig:humaneva_qualitative_results}
	\includegraphics[width=\textwidth]{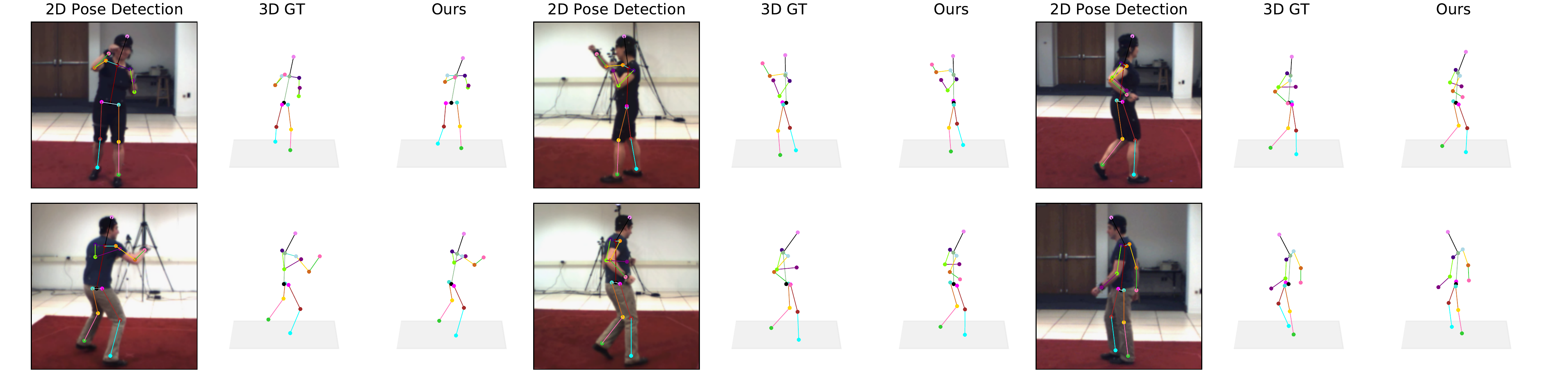}}
 
\subfloat[Qualitative results on the MPII-INF-3DHP dataset. Without fine-tuning the model can generalize to the challenging MPII-INF-3DHP dataset, which contains scenes recorded in-the-wild, and sufficiently predicted the 3D body pose.]{
	\label{subfig:3dhp_qualitative_results}
	\includegraphics[width=\textwidth]{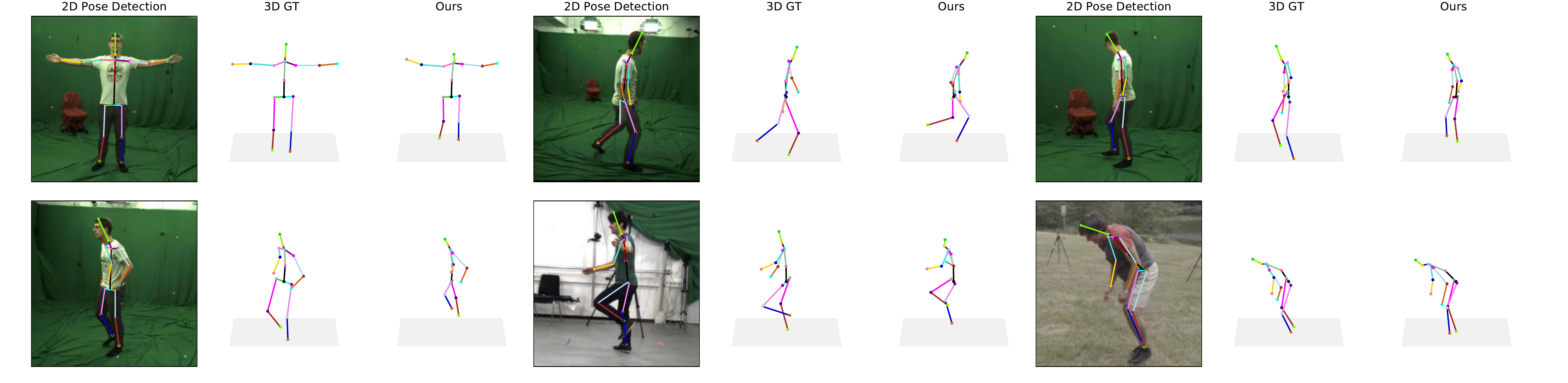}} 
\endgroup
 
\caption{Qualitative results on the three benchmarks, Human3.6M~\cite{ionescu2013human3}, HumanEVA~\cite{sigal2010humaneva} and MPII-INF-3DHP~\cite{mpiiinfdataset}. From left to right we show the detected 2D pose on the image, the 3D body pose ground-truth (3D GT) and the the 3D prediction of our model. The model can generalize to unseen scenes and subjects.}
\label{fig:qualitative_results}
 
\end{figure*}

HumanEva~\cite{sigal2010humaneva} has three subjects recorded from three camera views at 60 Hz. For our evaluation, we follow the protocol from~\cite{Martinez17}. We adopt the 15-joint skeleton model and make use of the provided train and test splits. Inline with prior works \cite{Martinez17}, we train on all actions and subject and evaluate on \textit{Walking} and \textit{Jogging}. The evaluation metric is the mean per joint position error after the Procrustes alignment~\cite{Martinez17}, similar to Human3.6M evaluation. Table~\ref{tab:HumanEva} summarizes our results. Due to the lack of unsupervised learning approaches (as well as weakly supervised ones), we compare our results with supervised learning methods. We rank somewhere in the middle of supervised learning approaches without using a single annotation during training. Finally, we present qualitative evaluation in Fig.~\ref{subfig:humaneva_qualitative_results}. It can be seen that we make correct predictions even for occluded body joints.

\begin{table}[h]
\centering
\small
\hspace{-1mm}
\tabcolsep=0.9mm
\caption{Results on HumanEva~\cite{sigal2010humaneva} and comparison with previous supervised approaches. Our model outperforms some of the supervised methods.}
\begin{adjustbox}{max width=\linewidth}
\begin{tabular}{@{}l |lll |lll |l @{}}
\toprule
Method & \multicolumn{3}{c}{Walking} & \multicolumn{3}{c}{Jogging} &\\
& S1 & S3 & S3 & S1 & S2 & S3 & Avg\\
\midrule
Radwan~\etal~\cite{radwan2013monocular}   & 75.1 & 99.8 & 93.8 & 79.2 & 89.8 & 99.4   & 89.5 \\
Wang~\etal~\cite{wang2014robust}          & 71.9 & 75.7 & 85.3 & 62.6 & 77.7 & 54.4 &  71.3\\
Simo-Serra~\etal~\cite{simo2013joint}     & 65.1 & 48.6 & 73.5 & 74.2 & 46.6 & 32.2  & 56.7\\
Kostrikov~\etal~\cite{kostrikov2014depth} & 44.0 & 30.9 & 41.7 & 57.2 & 35.0 & 33.3   & 40.3\\
Yasin~\etal~\cite{Yasin2016}          & 35.8 & 32.4 & 41.6 & 46.6 & 41.4 & 35.4 & 38.9  \\
Pavlakos~\etal~\cite{pavlakos2017coarse}          & 22.1 & 21.9 & 29.0 & 29.8 & 23.6 & 26.0 & 25.5  \\
Martinez~\etal~\cite{Martinez17}                   & 19.7& 17.4 & 46.8 & 26.9& 18.2 & 18.6 &  24.6\\
\midrule 
Ours (self-supervised) & 59.2 & 60.3 & 52.2 & 38.2 & 61.7 & 81.3 &  \textbf{58.9}\\
\bottomrule
\end{tabular}
\end{adjustbox}
\vspace{0.1mm}
\label{tab:HumanEva}
\end{table}

\subsection{Ablation Study}
We evaluate each loss function to show how the P-MPJPE metric is reduced by incrementally adding all loss functions of our algorithm. The results are reported in Table \ref{tab:ablationstudies} where the order of including additional loss functions is the same as the presentation of our loss functions in Sec.~\ref{loss_func_sec}.

We obtain already a competitive result using only the input triangulation loss. By incrementally adding the rest loss functions, we can improve our approach by $8\%$. We provide qualitative results in Fig.~\ref{subfig:h36m_qualitative_results}.

\begin{table}[h]
\centering
\small
\hspace{-3mm}
\tabcolsep=3.0mm
\caption{Ablation Study Results. The loss ablations show the effect of each loss function on the final model performance. We incrementally add each of the proposed loss functions.}
\begin{tabular}{@{}l |lll |lll l @{}}
\toprule
Loss Functions & P-MPJPE\\
\midrule
$\mathcal{L}_{in}$   & 67.5 \\
$\mathcal{L}_{proj}$   & 68.0 \\
$\mathcal{L}_{in}$ +  $\mathcal{L}_{proj}$  & 65.3\\
$\mathcal{L}_{in}$ +  $\mathcal{L}_{proj}$ + $\mathcal{L}_{con}$  & 63.9\\
$\mathcal{L}_{in}$ +  $\mathcal{L}_{proj}$ + $\mathcal{L}_{con}$ + $\mathcal{L}_{out}$ & \textbf{62.0}\\
\bottomrule
\end{tabular}
\label{tab:ablationstudies}
\end{table}

\subsection{Generalization Evaluation}

In this experiment, we examine the generalization of our approach by testing a pre-trained model on Human3.6M on a different configuration without training on it. We assume that our method can generalize to another database since it does not directly depend on the image data or the calibration system. Nevertheless, we are obliged to the object detection performance, which affects the performance of our approach. As test set, we consider the MPII-INF-3DHP~\cite{mpiiinfdataset} dataset, which has 4 male and 4 female professional actors performing different actions, including diverse clothing and viewpoints. In our experiments, we rely only on the test set that has 2935 frames from 6 subjects performing 7 actions. We quantitatively evaluate our model pre-trained on Human3.6M to unseen scenes and subjects, and report the results in Table~\ref{tab:mpii_inf_test}. Our results are reported in 3D Percentage of Correct Keypoints (3DPCK) and the corresponding Area Under the Curve (AUC), which compose the standard metrics for the benchmark.

We compare our method with related approaches from supervised and weakly supervised learning, as shown in Table~\ref{tab:mpii_inf_test}. All approaches have been trained with the Human3.6M database. We achieve the best performance for supervised and self-supervised learning, while we have competitive results in weakly supervised learning. Our promising performance without extra training makes our approach suitable for skeleton-based input to tasks such as gesture recognition~\cite{wiederer2020traffic} or trajectory estimation~\cite{hasan2019forecasting}. We provide some visual results in Fig.~\ref{subfig:3dhp_qualitative_results}.

\begin{table}[t!]
\centering
\small
\tabcolsep=3.0mm
\caption{Our results on MPII-INF-3DHPE. All approaches are trained on Human3.6M. Our method outperforms all self-supervised methods trained without 3D ground-truth information and achieves similar performance to supervised and weakly-supervised approaches in the transfer task.}
\begin{adjustbox}{max width=\linewidth}
\begin{tabular}{@{}l l|ll lll l @{}}
\toprule
Supervision & Method & Trainset & PCK & AUC \\
\midrule
Supervised   & Mehta~\cite{mpiiinfdataset} & H36M & 64.7 & 31.7\\
Weakly supervised   & Zhou  \cite{zhou2016sparseness} & H36M & 69.2  & 32.5\\
\midrule
\multirow{2}{*}{Self-supervised}  & Chen et al. \cite{chenunsupervised19} & H36M & 64.3  & 31.6\\
            & Ours & H36M & \textbf{65.9} & \textbf{32.5} \\
\bottomrule
\end{tabular}
\end{adjustbox}
\label{tab:mpii_inf_test}
\end{table}

\section{Conclusion}

We presented a self-supervised learning algorithm for 3D human pose estimation, which is based on a multiple-view camera system and 2D body pose estimates for each view. To train our model, represented by a deep neural network, we propose a four-loss function training algorithm that does not require any kind of 2D or 3D body pose annotation. We evaluated our approach on Human3.6M and HumanEva databases, the standard benchmarks for 3D human pose estimation. Finally, we further examined the generalization of our approach on a different environment and also analysed the contribution of our loss functions. Overall, our method achieves state-of-the-art performance in all evaluations when compared to self-supervised learning approaches without 3D ground-truth. Besides, we reach comparable results to supervised and weakly-supervised learning approaches.

\section{ACKNOWLEDGMENTS}
Part of this work was supported by the research project "KI Delta Learning" (project number: 19A19013A) funded by the Federal Ministry for Economic Affairs and Energy (BMWi) on the basis of a decision by the German Bundestag.


{\small
\bibliographystyle{ieee}
\bibliography{egbib}
}

\end{document}